\documentclass[final]{cvpr}

\usepackage{times}
\usepackage{epsfig}
\usepackage{graphicx}
\usepackage{amsmath}
\usepackage{amssymb}
\usepackage{subcaption}
\usepackage{booktabs}
\usepackage{hyperref}
\usepackage{placeins}
\usepackage{dsfont}
\usepackage{animate}

\newcommand\blfootnote[1]{%
  \begingroup
  \renewcommand\thefootnote{}\footnote{#1}%
  \addtocounter{footnote}{-1}%
  \endgroup
}
\usepackage{pifont}
\DeclareUnicodeCharacter{2014}{\dash}

\def\cvprPaperID{1713} 
\def\confYear{CVPR 2021}
\begin{document}

\title{From Goals, Waypoints \& Paths To Long Term Human Trajectory Forecasting}
	
\author{Karttikeya Mangalam$^{\dagger*}$ \;\; Yang An$^{\S*}$ \;\; Harshayu Girase$^{\dagger}$ \;\; Jitendra Malik$^{\dagger}$ \;\; \vspace{10pt}\\ 
	$^\dagger$ UC Berkeley  \;\;  
	$^\S$ Technical University of Munich \;\;  
	}

\twocolumn[{
\renewcommand\twocolumn[1][]{#1}
\maketitle
\begin{center}
\centering
    \centering
    \includegraphics[width=0.19\textwidth]{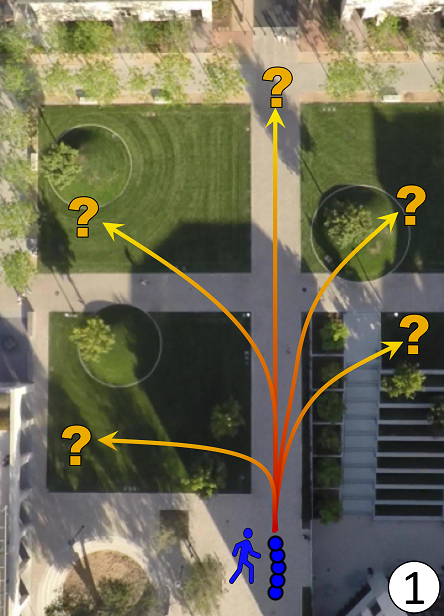}
    \includegraphics[width=0.19\textwidth]{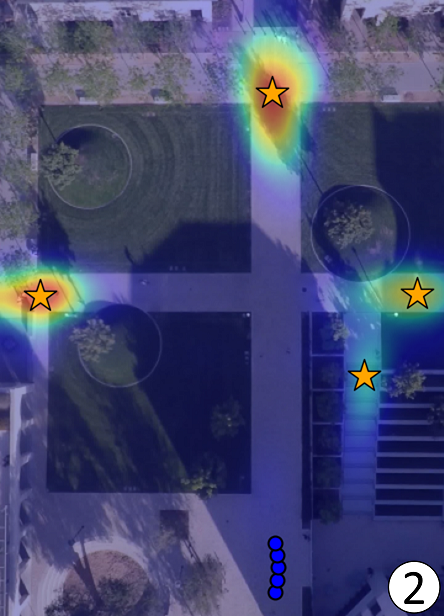}
    \includegraphics[width=0.19\textwidth]{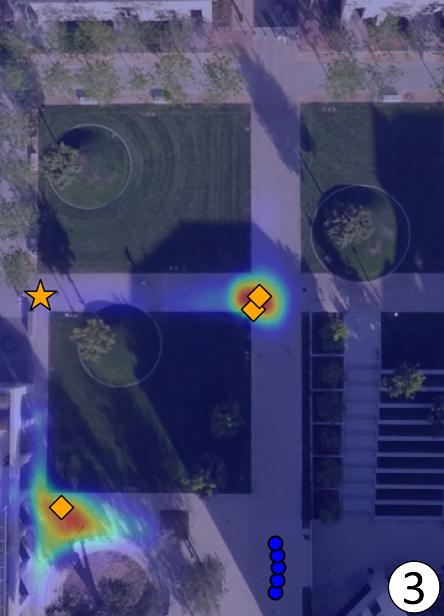}
    \includegraphics[width=0.19\textwidth]{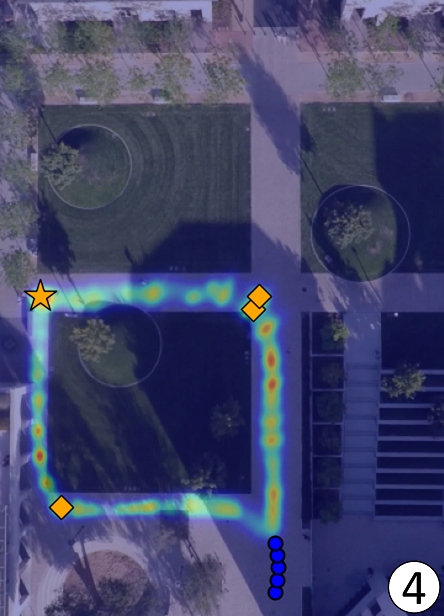}
    \includegraphics[width=0.19\textwidth]{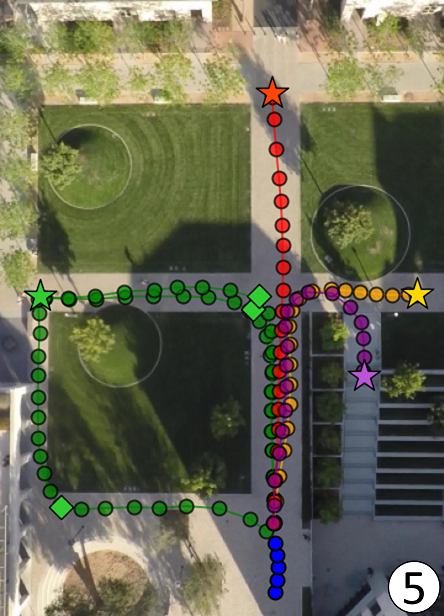}
   \captionof{figure}{We tackle the problem of long term human trajectory forecasting. Given the past motion of an agent (blue) on a scene over the last five seconds, we aim to predict the multimodal future motion upto the next minute \ding{192}. To achieve this, we propose to factorizing overall multimodality into its \textit{epistemic} and \textit{aleatoric} factors. The \textit{epistemic} factor is modeled with an estimated distribution over the long term goals \ding{193} while the \textit{aleatoric} factor is modeled as a distribution over the intermediate waypoints \ding{194} and trajectory \ding{195}. This is repeated for multiple goals and waypoints in parallel for scene-compliant multimodal human trajectory forecasting \ding{196}. Each color indicates predicted trajectories for different sampled goals.}
    \label{fig:teaser}
\end{center}
}]

\begin{abstract}
\vspace{-8mm}
Human trajectory forecasting is an inherently multimodal problem. Uncertainty in future trajectories stems from two sources: (a) sources that are known to the agent but unknown to the model, such as long term goals and (b) sources that are unknown to both the agent \& the model, such as intent of other agents \& irreducible randomness in decisions. We propose to factorize this uncertainty into its \textit{epistemic} \& \textit{aleatoric} sources. We model the \textit{epistemic} uncertainty through multimodality in long term goals and the \textit{aleatoric} uncertainty through multimodality in waypoints \& paths. To exemplify this dichotomy, we also propose a novel long term trajectory forecasting setting, with prediction horizons upto a minute, an order of magnitude longer than prior works. Finally, we present $\textsf{Y}$-net, a scene compliant trajectory forecasting network that exploits the proposed epistemic \& aleatoric structure for diverse trajectory predictions across long prediction horizons. $\textsf{Y}$-net significantly improves previous state-of-the-art performance on both (a) The well studied short prediction horizon settings on the Stanford Drone \& ETH/UCY datasets and (b) The proposed long prediction horizon setting on the re-purposed Stanford Drone \& Intersection Drone datasets.
\end{abstract}

\vspace{-6mm}
\section{Introduction}
\label{sec:intro}

Sequence prediction is a fundamental problem in several engineering disciplines such as signal processing, pattern recognition, control engineering and in virtually any domain concerned with temporal measurements. From the seminal work of A. A. Markov \cite{aamarkov} on predicting the next syllable in the poem \textit{Eugene Onegin} with Markov chains, to modern day autoregressive descendants like GPT-3 \cite{brown2020language}, next element prediction in a sequence has a long standing history. Time series forecasting is a key instantiation of the sequence prediction problem in the setting where the sequence is formed by elements sampled in time. Several classic techniques such as Autoregressive Moving Average Models (ARMA) \cite{whittle1951hypothesis} have been incorporated in deep learning architectures \cite{vasishta2017natural, hochreiter1997long} in modern day state-of-the-art time series forecasting methods \cite{sen2019think}. 

\blfootnote{$*$ indicates equal contribution.}

However, humans are not inanimate Newtonian entities, slave to predetermined physical laws \& forces. Predicting the future motion of a billiard ball smoothly rolling on a pool table under friction and physical constraints is a problem of different nature from forecasting human motion and positions. Humans are goal conditioned agents that, unlike the ball, exert their will through actions to achieve a desired outcome \cite{tomasello2005understanding}. Anticipating human motion is of fundamental importance to dynamic agents such as other humans, autonomous robots \cite{bennewitz2002learning} and self-driving vehicles \cite{thrun2002probabilistic}. Human motion is inherently goal directed and is put in place by the agent to bring about a desired effect.  

Nevertheless, even conditioned on the agent's past motion and overarching long term goals, is the future trajectory deterministic? Consider yourself standing at a crossing on a busy street, waiting for the pedestrian light to turn green. While you have every intention of crossing the street, the exact future trajectory remains stochastic as you might swerve to avoid other pedestrians, speed up your pace if the light is about to turn red, or pause abruptly if an unruly cyclist dashes by. Hence, even conditioned on the past observed motion and scene semantics, future human motion is inherently stochastic \cite{helbing1993stochastische} owing to both \textit{epistemic} uncertainty caused by latent decision variables like long term goals and \textit{aleatoric} variability \cite{der2009aleatory} stemming from random decision variables such as environmental factors. This dichotomy is even sharper in long term forecasting, since due to the increased uncertainty in future, the aleatoric randomness influences the trajectory much more strongly in long rather than short temporal horizons. 

This motivates a factorized multimodal approach for human dynamics modeling where both factors of stochasticity are modeled hierarchically rather than lumped jointly. We hypothesize that the long term latent goals of the agent represent the \textit{epistemic} uncertainty with motion prediction. This is motivated by the observation that while the agent has a goal in mind while planning and executing their trajectory, this is unknown to the prediction system. In physical terms, this is akin to the question of \textit{where} the agent wants to go. Similarly, the \textit{aleatoric} uncertainty is expressed in the stochasticity of the path leading to the goal, which encompasses factors like agent's handedness, environment variables such as other agents, partial scene information available to the agent and most importantly, the unconscious randomness in human decisions \cite{kahneman2011thinking}. In physical terms, this is akin to the question of \textit{how} the agent reaches the goal.

Hence, we propose to model the \textit{epistemic} uncertainty first and then the \textit{aleatoric} stochasticity conditioned on the obtained estimate. Concretely, with the RGB scene and the past motion history we first estimate an explicit probability distribution over the agent's final positions at the end of the trajectory, \ie the agent's long term goals. This represents the \textit{epistemic} uncertainty in the prediction system. We also estimate distributions over a few chosen future waypoint positions which along with the sampled goal points are used to obtain explicit probability maps over all the remaining trajectory positions. This represents the aleatoric uncertainty in the prediction system. Together the samples from the \textit{epistemic} goal and the \textit{aleatoric} waypoint \& trajectory distribution form the predicted future trajectory.    

In summary, our contribution is threefold. \textbf{First}, we propose a novel long term prediction setting that extends upto a minute in the future which is about an order of magnitude longer than previous literature. We also benchmark performance of previous state-of-the-art short horizon prediction models on this setting along with simple baselines. \textbf{Second}, we propose $\textsf{Y}$-net, a scene-compliant long term trajectory prediction network that explicitly models both the \textit{goal} and \textit{path} multimodalities by making effective use of the scene semantics. \textbf{Third}, we show that the factorized multimodality modeling enables $\textsf{Y}$-net to improve the state-of-the-art both on the proposed long term settings and the well-studied short term prediction settings. We benchmark $\textsf{Y}$-net's performance on the Stanford Drone \cite{robicquetlearning} and the ETH \cite{pellegrini2010improving}/UCY\cite{lerner2007crowds} benchmark in the short term setting, where it outperforms previous approaches by significant margins of $26.9\%$ \& $5.6\%$ respectively, on ADE and by $34.0\%$ and $51.9\%$ respectively, on FDE metric. Further, we also study $\textsf{Y}$-net's performance in the proposed long term prediction setting on the Stanford Drone \& the Intersection Drone Dataset \cite{inDdataset} where it substantially improves the performance of state-of-the-art short term methods by over $50.6\%$ and $35.0\%$ respectively, on ADE and $77.1\%$ and $55.9\%$ respectively, on FDE metric.  

\section{Related Works}
Several recent studies have investigated human trajectory prediction in different settings. Broadly, these approaches can be grouped on the basis of the proposed formulation for multimodality in forecasting, inputs signals available to the prediction model and the nature and form of prediction results furnished by the model. Several diverse input signals such as agent's past motion history \cite{helbing1995social}, human pose \cite{mangalam2020disentangling}, RGB scene image \cite{gupta2018social, sadeghian2019sophie, caoHMP2020, lee2017desire, liang2020garden}, scene semantic cues \cite{caoHMP2020}, location \cite{salzmann2020trajectron++, li2019conditional, bhattacharyya2019conditional}  \& gaze of other pedestrian \cite{mangalam2020disentangling, yagi2018future} in the scene, moving vehicles such as cars \cite{salzmann2020trajectron++} and also latent inferred signals such as agent's goals \cite{mangalam2020not} have been used. The form of prediction results produced are also diverse with multimodality \cite{liang2020garden} and scene-compliant forecasting being central to the prior works. 

\subsection{Unimodal Forecasting}
Early trajectory forecasting work focused on unimodal predictions of the future. Social Forces \cite{helbing1995social} proposes modeling interactions as attractive and repulsive forces and future trajectory as a deterministic path evolving under these forces. Social LSTM \cite{alahi2016social} focuses on other agents in the scene and models their effects through a novel pooling module. \cite{yagi2018future} tackles motion forecasting in ego-centric views and develops a system that exploits subtle cues like body pose and gaze along with camera wearer's ego-motion for other agent's future location prediction. \cite{vemula2018social} proposes to use attention to model current agent's interaction with other agent's. \cite{mangalam2020disentangling} predicts trajectory as the `global' branch for pose prediction and proposes to condition downstream tasks such as pose prediction on predicted unimodal trajectories. 

\subsection{Multimodality through Generative Modeling}
A line of work aims to model the stochasticity inherent in future prediction through a latent variable with a defined prior distribution through approaches such as conditional variational auto-encoders \cite{kingma2013auto}. Lee \etal \cite{lee2017desire} propose DESIRE, an inverse reinforcement learning based approach that uses multimodality in sampling of a latent variable that is ranked and optimized with a refinement module. CF-VAE \cite{bhattacharyya2019conditional} uses normalizing flows with VAEs for modeling structure in sequences such as trajectories. \cite{mangalam2020disentangling} introduces the use of a CVAE for capturing multimodality in the final position of the pedestrians conditioned on the past motion history. Trajectron++ \cite{salzmann2020trajectron++} represents agent's trajectories in a graph structured recurrent network for scene complaint trajectroy forecasting, taking into account the interaction with a diverse set of agents. CGNS \cite{li2019conditional} uses variational divergence minimization procedure in multimodal latent space to learn feasible regions for future trajectories.

A different line of work includes Social GAN \cite{gupta2018social} which uses adversarial losses \cite{goodfellow2014generative} for incorporating multimodality in predictions. SoPhie \cite{sadeghian2019sophie} further incorporates attention modules to model agent's interactions with the environment and other agents.

While such generative approaches do produce diverse trajectories, overall coverage of critical modes cannot be guaranteed and little control is afforded over the properties of predicted trajectories such as direction, number of samples \etc. In contrast, our method, $\textsf{Y}$-net, estimates explicit probability maps which allow easily incorporating spatial constraints for a downstream task.  

\subsection{Multimodality through spatial probability estimates}
Another line of work obtains multimodality via estimated probability maps. Activity Forecasting from Kitani \etal \cite{kitani2012activity} proposes to use a hidden Markov Decision process for modeling the future paths. However, in contrast to our work the future predictions in \cite{kitani2012activity} are conditioned on the activity label such as `approach car', 'depart car' \etc. More recently, some works have used a grid based scene representation for estimating probabilities for future time steps \cite{liang2020simaug, liang2020garden, deo2020trajectory}. Relatedly, some prior works such as \cite{mangalam2020disentangling, zhao2020tnt, caoHMP2020} propose a goal-conditioned trajectory forecasting method. However, no prior works have proposed factorized modeling of \textit{epistemic} uncertainty or goals \& \textit{aleatoric} uncertainty or paths as $\textsf{Y}$-net uses. 

\begin{figure*}[t!]
\begin{center}
\includegraphics[width = \textwidth]{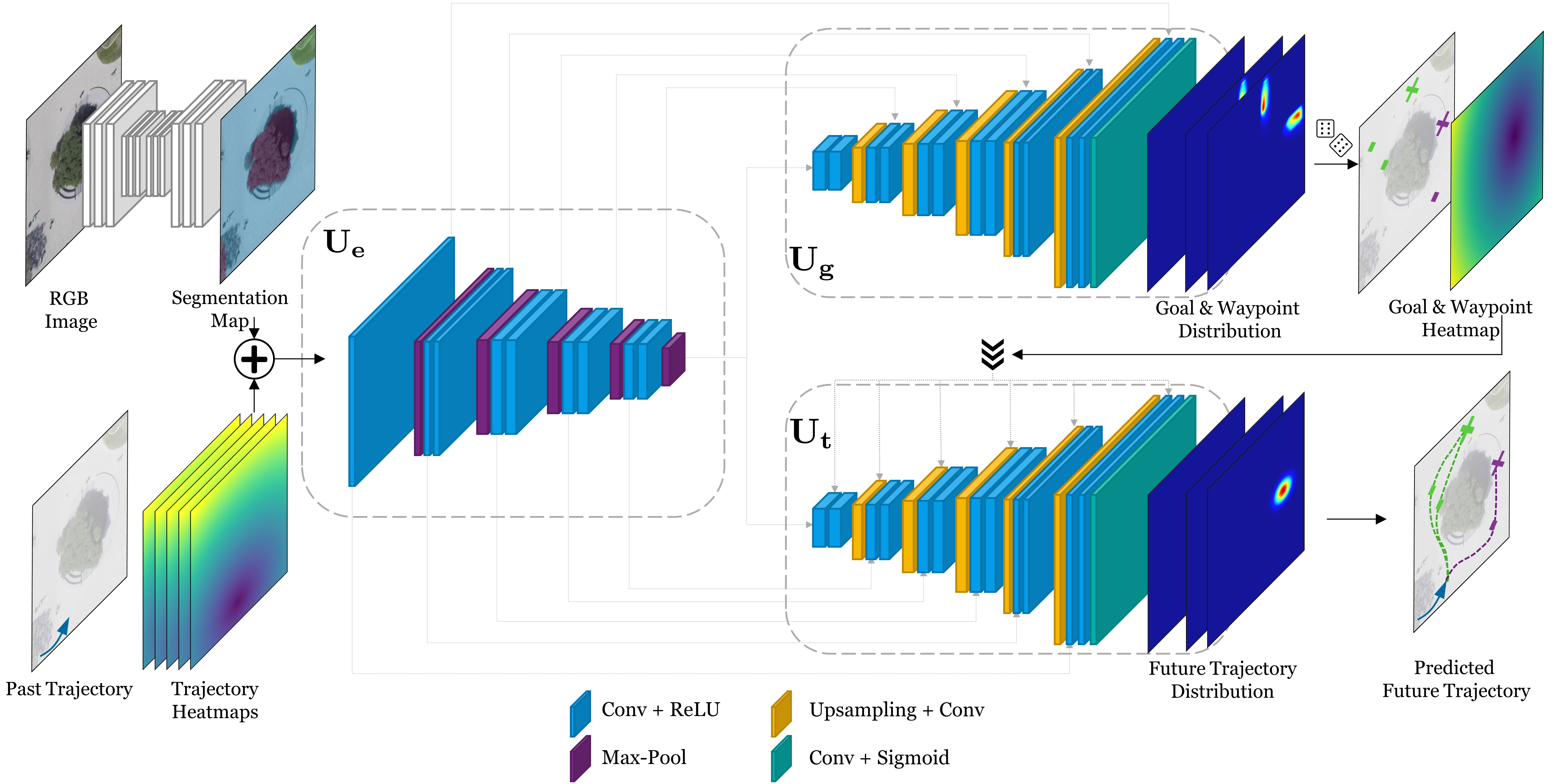}
\end{center}

  \caption{\textbf{Model Architecture}: \textsf{Y}-net comprises of three sub-networks $\mathbf{U}_e$, $\mathbf{U}_g$ \& $\mathbf{U}_t$ modeled after the Unet architecture \cite{unet} (Section \ref{sec:traj_scene_rep}). $\textsf{Y}$-net adopts a factorized approach to multimodality, expressing the stochasticity in goals \& waypoints through estimated distributions furnished by $\mathbf{U}_g$. And multimodality in paths is achieved through estimated probability distributions obtained by $\mathbf{U}_t$ conditioned on samples from $\mathbf{U}_g$ for predicting diverse multimodal scene-compliant futures.}
  
\label{fig:arch}
\end{figure*}

\section{Proposed Method}
The problem of multimodal trajectory prediction can be formulated formally as follows. Given a RGB scene image $\mathcal{I}$ and past positions of a pedestrian in the scene $\mathcal{I}$ denoted by $\{\mathbf{u}_n\}_{n = 1}^{n_p}$ for the past $t_p = n_p/\texttt{FPS}$ seconds sampled at the frame rate $\texttt{FPS}$, the model aims to predict the position of the pedestrian for the next $t_f$ seconds in the future, denoted by $\{\mathbf{u}^{i}_n\}_{n = n_p}^{n_p + n_f}$ where $t_f = n_f/\texttt{FPS}$. 

Since the future is stochastic, multiple predictions for the future trajectories are produced. In this work, we factorize the overall stochasticity into two modes. First are the modes relating to \textit{epistemic} uncertainty \ie multimodality in the final destination of the agent for which the module produces $K_e$ predictions. Second are the modes relating to the \textit{aleatoric} uncertainty \ie multimodality in the path taken to the destination stemming from uncontrolled randomness given the goal, for which the module produces $K_a$ separate predictions for each given destination. In the short temporal horizon limit, since the overall path length is small, the options for paths to a given goal are limited and similar to each other. Hence, this results in setting $K_a = 1$ and so the total paths predicted ($K$ in prior works) is the same as $K_e$. However, for longer temporal horizons, there are several paths to the same goal and hence $K_a > 1$. Next, we describe in detail the working of our model, $\textsf{Y}$-net and its three sub-networks $\mathbf{U}_e$, $\mathbf{U}_g$ and $\mathbf{U}_t$ (Section \ref{sec:traj_scene_rep}) followed by details of the non-parametric sampling process (Section \ref{sec:sampling}) and loss functions (Section \ref{sec:loss}) used.

\subsection{\textsf{Y}-net Sub-Networks}
\label{sec:traj_scene_rep}

To effectively use the scene information in the semantic space with the trajectory information expressed in coordinates, alignment needs to be created between the different signals. Some prior works \cite{sadeghian2019sophie} achieve this by encoding the two-dimensional RGB image $\mathcal{I}$ as a one-dimensional hidden state vector extracted from some pretrained network. While this provides the network with scene information, any meaningful spatial signal gets highly conflated when flattened into a vector and pixel alignment is destroyed. This is highlighted in \cite{mangalam2020not} which establish previous state-of-the-art without any RGB information underlining the misuse of image information in prior works. In this work, we adopt a trajectory on scene heatmap representation that solved the alignment issue by representing the trajectory spatially in the same two-dimensional space as image $\mathcal{I}$. 

\subsubsection{Trajectory on Scene Heatmap Representation}
\label{sec:traj_on_scene_subsec}
The RGB image $\mathcal{I}$ is first processed with a semantic segmentation network such as U-net \cite{unet} that produces segmentation map $\mathbf{S}$ of $\mathcal{I}$ comprising of $N_c$ classes determined according to the affordance provided by the surface to an agent for actions such as walking, standing, running etc. In a parallel branch, the past motion history $\{\mathbf{u}_n\}_{n = 1}^{n_p}$ of agent $p$ is converted to a trajectory heatmap $\mathbf{H}$ of spatial sizes of $\mathcal{I}$ and $n_p$ channels corresponding to the past $t_p$ seconds sampled at the frame rate. Mathematically,   
$$
\mathbf{H}(n,i,j) = 2 \frac{\Vert (i,j) - \mathbf{u}_{n} \Vert}{\max\limits_{{(x,y) \in \mathcal{I}}}\Vert (x,y) - \mathbf{u}_{n} \Vert}
$$
The heatmap trajectory representation is then concatenated with the semantic map $\mathcal{S}$ along the channel dimension producing the trajectory on scene heatmap tensor $\mathbf{H}_{\mathcal{S}}$ a ${H \times W \times (N_c + n_p)}$ dimensional input tensor which is passed to the encoder network $\mathbf{U}_e$.  

\subsubsection{Trajectory on Scene Heatmap Encoder}
\label{sec:encoder}
The tensor $\mathbf{H}_{\mathcal{S}}$ is processed with the encoder $\mathbf{U}_e$ designed as a U-net encoder \cite{unet} as shown in Figure \ref{fig:arch}. The encoder $\mathbf{U}_e$ consists of a total of $N_{\mathbf{U}_e}$ blocks, where the spatial dimensions are reduced from $H \times W$ to $H_{\mathbf{U}} \times W_{\mathbf{U}}$ halving after every block using max pooling and channel depth is increased sequentially from $N_c + n_p$ to $C_{\mathbf{U}_e}$ doubling after a certain number of blocks.
The final spatially compact and deep representation $\mathbf{H}_{\mathbf{U}_e}$ along with the $N_{\mathbf{U}_e} - 1$ intermediate feature tensors of varying spatial resolution are passed onto the goal decoder $\mathbf{U}_g$ and the trajectory decoder $\mathbf{U}_t$ as discussed below.

\subsubsection{Goal \& Waypoint Heatmap Decoder}
\label{sec:goal_decoder}
The feature maps of ${U}_e$ at various spatial resolutions are passed onto the goal \& waypoint heatmap decoder $\mathbf{U}_g$ which is modeled after the expansion arm in the U-net architecture \cite{unet}. After a center block consisting of two convolutional layers which takes the most compact feature map $\mathbf{H}_{\mathbf{U}_e}$ from $\mathbf{U}_e$, each block of the expansion arm starts by expanding the previous feature map, spatially doubling the resolution in every block using bilinear up-sampling and convolution (together forming Deconvolution \cite{unet}). 

Further, after every Deconvolution, the corresponding intermediate representations from $\mathbf{U}_e$ are merged and the features are passed through two convolution layers.
Merging intermediate high resolution feature maps from $\mathbf{U}_e$ is necessary since just using $\mathbf{H}_{\mathbf{U}_e}$ would severely limit the final resolution of the goal heatmap, thus missing fine spatial details that are preserved in the intermediate feature maps. 

Deconvolution, feature merging and convolution form a U-net block. $\mathbf{U}_g$ consists of $N_{\mathbf{U}_g}$ blocks, followed by a per-pixel sigmoid that produces $N^w + 1$ estimated spatial heatmaps. This includes the probability distribution of the final goal of the agent \ie the non-parametric distribution $\hat{\mathbb{P}}(\mathbf{u}_{n_p + n_f})$ and $N^w$ intermediate waypoint probability heatmaps at frame steps ${w_i} \in \{n_p, n_p+1, \hdots ,n_p + n_f\} \ \forall\  i = 1, \hdots, N^w$ represented as non-paramateric distributions $\hat{\mathbb{P}}(\mathbf{u}_{w_i})$.

\subsubsection{Trajectory Heatmap Decoder}
\label{sec:traj_decoder}
The estimated goal and waypoint distributions are sampled as described in Section \ref{sec:sampling} and the obtained sample $\mathbf{\hat{u}}_{n_p + n_f}$ for the goal and $\{\mathbf{\hat{u}}_{w_i}\}_{i=1}^{N^w}$ for the intermediate waypoints are converted to a heatmap representation as described in Section \ref{sec:traj_scene_rep}. The obtained conditioning tensor $\mathbf{H}_{\mathbf{U}_g}$ is spatially downsampled to match the corresponding block's size. Those are passed along with the past motion and scene representation $\mathbf{H}_{\mathbf{U}_e}$ and the $N_{\mathbf{U}_e} - 1$ intermediate high resolution tensors to the trajectory decoder network $\mathbf{U}_t$. $\mathbf{U}_t$ is modeled after the expansion arm of a U-net as well and proceeds in a similar fashion as $\mathbf{U}_g$, as described in Section \ref{sec:goal_decoder} with a total of $N_{\mathbf{U}_t}$ expansion blocks. The final trajectory distribution is obtained with a channel independent per-pixel sigmoid that produces for each future timestep a separate heatmaps of spatial size $H \times W$ corresponding to the position of the agent in the scene over the next $n_f$.

\subsection{Non-parametric Distribution Sampling}
\label{sec:sampling}
Given a distribution $\mathbb{P}$ of the position of the agent in a future frame represented non-parametrically as a matrix of probabilities $X$, 
we aim to sample a two-dimensional point as our estimate for the position of the agent at that time step. This is difficult to achieve reliably in practice since the estimated distribution $\mathbb{P}$ is noisy. Hence, taking a naive $\texttt{argmax}$ is not robust. Instead we propose to use the $\texttt{softargmax}$ operation \cite{goodfellow20166} to approximate the most likely position in a robust \& stable fashion. Mathematically,  
\begin{equation*}
\texttt{\small{softargmax}}(X) = \left(\sum_{i} i \frac{\sum_j e^{X_{ij}}}{\sum_{i,j} e^{X_{ij}}}, \sum_{j} j \frac{ \sum_i e^{X_{ij}}}{\sum_{i,j} e^{X_{ij}}}\right) 
\end{equation*}

\subsubsection{Test-Time Sampling Trick ($\texttt{TTST}$)} 
In situations where multiple samples are required, such as during testing, sampling from $\mathbb{P}$ can be carried in a straightforward fashion by considering $X$ as a categorical distribution with given probabilities $X(i,j)$ for position $(i,j)$. However, this approach doesn't take into account the number of samples required from $\mathbb{P}$ all of which jointly will represent the quality of the estimated $\mathbb{P}$. For example, if only a few samples are required, sampling indiscriminately spatially is sub-optimal since samples are likely to be wasted if drawn from low probability regions or sampled from adjacent regions.

We propose a `Test-Time Sampling Trick' ($\texttt{TTST}$) that is cognizant of the number of goal samples, $K_e$, needed for evaluation. During testing, we propose to first sample a large number of points (10,000 in our experiments) from the estimated distribution $\mathbb{P}$ in a $K_e$-agnostic manner. 

To eliminate outliers, we suppress samples from pixels $(i,j)$ with probability $X_{ij}$ below a threshold $thr_{rel}$. It is set adaptively for each probability matrix $X$ separately to a fraction of the highest occurring value in that matrix: $thr_{rel}=\max (X) * 0.01$. 

Further, to control the tradeoff between diversity and precision, we use the hyper parameter temperature $T$. Before the pixel-wise sigmoid operation, we divide the predicted logit probability map through $T$. Lower temperature values results in probability maps $X$ with low entropy, \ie the probability mass is concentrated in a smaller number of pixels and samples are increasingly drawn from more likely regions. Higher temperature values increase the diversity of samples. For the short term setting, we use $T=1.0$, while for our proposed long term setting, we increase the diversity by setting $T=1.8$.

Then, we propose to run the fast clustering algorithm K-means on these sampled points with the number of clusters set to $K_e-1$. The cluster centers obtained from the K-means algorithm, along with the $\texttt{softargmax}$ sampled point,  form the final set of $K$ samples to be used for evaluation. Note that while in spirit this is similar to the `truncation trick' proposed in PECNet \cite{mangalam2020not}, the `truncation trick' is K-agnostic  and requires a well suited $\sigma_T$ to be chosen experimentally beforehand for a given $K$. Further, their `truncation trick' operates in the latent variable space with no direct control over the final generated samples since in \cite{mangalam2020not} multi-modality is introduced through implicit approaches like Variational Auto-encoders. Alleviating these limitations, $\texttt{TTST}$ provides direct control on the sampled points, is cognizant of the number of samples needed $K$ and hence, does not require any $K$-specific tuning because of the design choice of using explicit probability heatmaps. 

\begin{figure*}[h!]
\centering
    \includegraphics[height=0.24\textwidth, angle=90,trim={1cm 0 3cm 0}, clip]{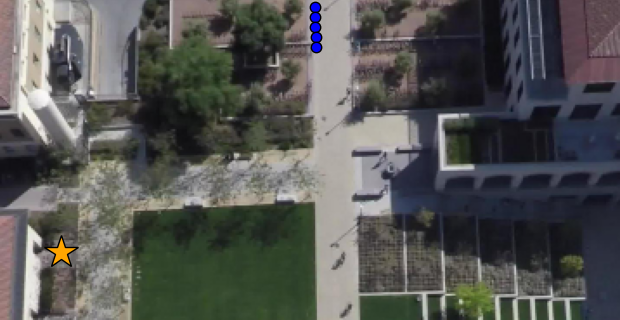}
    \includegraphics[height=0.24\textwidth, angle=90,trim={1cm 0 3cm 0}, clip]{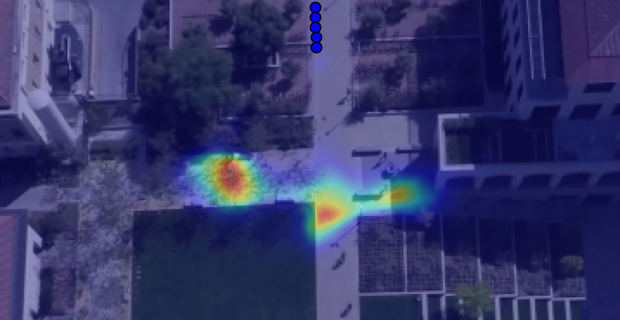}
    \includegraphics[height=0.24\textwidth, angle=90,trim={1cm 0 3cm 0}, clip]{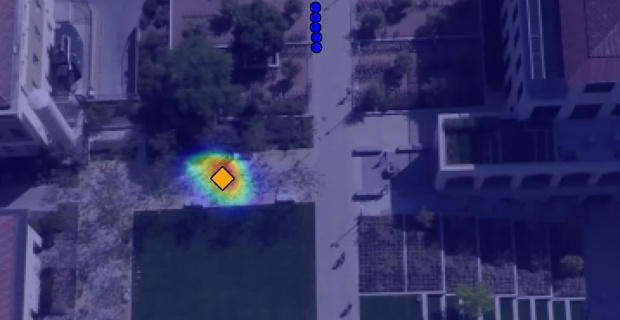}
    \includegraphics[height=0.24\textwidth, angle=90,trim={1cm 0 3cm 0}, clip]{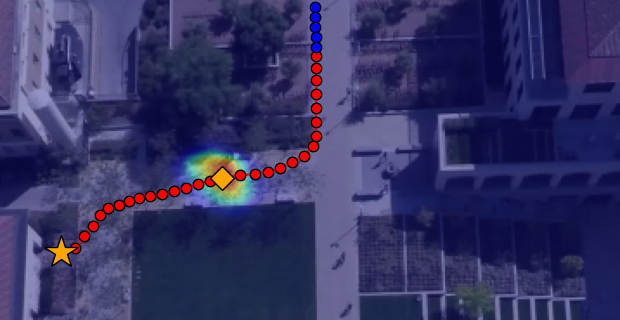}
   \caption[short]{\textbf{Conditioned Waypoint sampling}: The first image shows the scene with the past trajectory in blue. The yellow star indicates the sampled goal. The unconditioned waypoint distribution can be seen in the second image; the distribution is goal-agnostic and therefore probability mass is distributed over all three roads. The third image is the resulting waypoint distribution, by multiplying the multivariate Gaussian prior with the unconditioned prediction. The final image shows the trajectory towards the sampled goal while crossing the waypoint.}
\label{fig:waypoint}
\end{figure*}

\subsubsection{Conditioned Waypoint sampling}
Goal and waypoints are dependent to each other. Figure \ref{fig:waypoint} shows an example, where the road forks into three different paths. If the sampled goal lies on the bottom path, the agent most likely won't pass through a waypoint on the upper or middle path, and will prefer a waypoint on the bottom road.

Following this intuition, we introduce a hierarchical prior to condition the waypoint sampling on the already sampled goal and waypoints.

We first sample the goal. By fixing the goal the possible locations of the next waypoint is constrained. We assume that the waypoint lies on a straight line segment between the sampled goal and the past trajectory at $\frac{w_i-n_p}{n_f}$ of the line segment's length, \eg with $N^w=1$ waypoint, it would lie in the middle of the segment. This assumption is too hard and can lead to unrealistic paths not complying with the environment constraints. To relax this assumption, we use a multivariate Gaussian prior with mean at the assumed location. The variance is chosen adaptively by considering the distance between the agent's last observed position and the sampled goal as $\sigma_{\bot}=||u_{n_p+n_f}-u_{n_p}||/\alpha$ where $\alpha$ is a scaling hyper parameter. We set $\alpha=6$ in our experiments. Intuitively, the greater the distance between the current position and the sampled goal, the more uncertain is the waypoint position. $\sigma_{\bot}$ is the variance perpendicular to the line segment, and we set the variance parallel to the line segment to $\sigma_{\parallel}=\beta*\sigma_{\bot}$ with $\beta=0.5$ in our experiments. This constrains the possible waypoint position more in the direction of travel and leaves more room for uncertainty in the perpendicular direction.

We multiply the described multivariate Gaussian prior pixel-wise to the predicted waypoint distribution. Fusing prior and predicted distribution leads to scene-compliant waypoints in the correct direction. As seen in the example, it suppresses probability mass on the upper and middle road. From the resulting distribution we use $\texttt{softargmax}$ for the first waypoint and sample the remaining $K_a-1$ waypoints randomly to get two-dimensional points from the distribution.

If there is more than one waypoint, \ie $N^w>1$, we repeat the above process for the next waypoint at $w_{i}$ and condition it to the previously sampled waypoint at $w_{i+1}$.

\subsection{Loss Function}
\label{sec:loss}
Since we use the trajectory on scene representation (Section \ref{sec:traj_scene_rep}) we impose losses directly on the estimated distribution $\hat{\mathbb{P}}$ rather than on the samples. The ground truth future is represented as a Gaussian heatmap $\mathbb{P}$ centered at the observed points with a predetermined variance $\sigma_H$. All three networks, $\mathbf{U}_e$, $\mathbf{U}_g$ and $\mathbf{U}_t$ are trained end to end jointly using a weighted combination of binary cross entropy losses on the predicted goal, waypoint and trajectory distributions. 

\begin{align*}
\mathcal{L}_{\text{goal}} &= \text{BCE}(\mathbb{P}(\mathbf{u}_{n_p + n_f}), \hat{\mathbb{P}}(\mathbf{u}_{n_p + n_f}) )
\\
\mathcal{L}_{\text{waypoint}} &= \sum_{i=1}^{N^w} \text{BCE}(\mathbb{P}(\mathbf{u}_{w_i}), \hat{\mathbb{P}}(\mathbf{u}_{w_i}))  \\
\mathcal{L}_{\text{trajectory}} &= \sum_{i=n_p}^{n_p + n_f} \text{BCE}(\mathbb{P}(\mathbf{u}_{i}), \hat{\mathbb{P}}(\mathbf{u}_{i})) \\
\mathcal{L} &= \mathcal{L}_{\text{goal}} + \lambda_1 \mathcal{L}_{\text{waypoint}} + \lambda_2 \mathcal{L}_{\text{trajectory}}
\end{align*}

\section{Results}
\subsection{Datasets}
We use a total of three datasets to study $\textsf{Y}$-net's performance -- the Stanford Drone Dataset (SDD) \cite{robicquet2016learning} (for both short \& long term), the Intersection Drone Dataset (InD) \cite{inDdataset} (for long term only) and the ETH \cite{pellegrini2010improving} / UCY \cite{lerner2007crowds} forecasting benchmark (for short term only).

\noindent \textbf{Stanford Drone Dataset}: We benchmark our proposed model on the popular Stanford drone dataset
\cite{robicquet2016learning} where several recently proposed methods have improved state-of-the-art performance significantly in the past few years \cite{PWCsdd}. The dataset is comprised of more than $11,000$ unique pedestrians across $20$ top-down scenes captured on the Stanford university campus in bird's eye view using a flying drone. It has over $40,000$ agent-scene interactions and has enjoyed very popular usage in trajectory prediction literature in the short temporal horizon setting.
For short term prediction, we follow the well-established preprocessed data and splits from the TrajNet benchmark \cite{sadeghian2018trajnet}  setup used by \cite{gupta2018social,sadeghian2019sophie, mangalam2020not, deo2020trajectory}, sampling at $\texttt{FPS} = 2.5$ yielding an input sequence of length $n_p = 8$ ($3.2$ seconds) and output of length $n_f = 12$ ($4.8$ seconds).

In our proposed long term setting, we split the raw data of Stanford Drone Dataset (SDD) in the same fashion as proposed in TrajNet benchmark \cite{sadeghian2018trajnet} evaluating on the same scenes, all of which are not seen during training. The raw data is recorded in $\texttt{FPS} = 30$ and we first downsample the data to our proposed $\texttt{FPS} = 1$, thus yielding a $n_p = 5$ for $t_f = 5$ seconds in the past and predicting upto one minute into the future. We use the middle point of the raw bounding boxes to get the same coordinate representation as the preprocessed short term setting. The data contains various types of agent beyond pedestrians (bicyclists, skateboarders, cars, buses, and golf carts), we filter out all non-pedestrians and short trajectories below $n_p+n_f$ out. As the raw data is noisy and contains temporal discontinuities, we split the trajectories at those discontinuities. We use a sliding window approach without overlap to split up long trajectories, resulting in our final dataset. After those steps, the dataset contains 1502 trajectories for $n_p=5$ and $n_f=30$. Further, we label the scenes with semantic segmentation maps consisting of the following $N_c = 5$ ``stuff" classes \cite{caesar2018coco} depending on the affordability of class to pedestrians: pavement, terrain, structure, tree and road.

\noindent \textbf{Intersection Drone Dataset}: We propose to use the Intersection drone dataset \cite{inDdataset} for benchmarking trajectory forecasting in long horizon settings. The dataset comprises over $10$ hours of measurements over $4$ distinct intersection in an urban environment. We use similar steps for the Intersection Drone Dataset as for SDD. To evaluate $\textsf{Y}$-net and the baselines performance on unseen scenes during training, we only use location ID 4 during testing. The raw video and detection are in $\texttt{FPS} = 25$, and again, we downsample the data to $\texttt{FPS} = 1$. We then filter out non-pedestrians and  short  trajectories  and  use  a  sliding  window  approach without overlap to split long trajectories. Since the data lies in world coordinates, we convert it into pixel coordinates by scaling with the provided scale factors from the authors. After the preprocessing steps, inD contains 1,396 long term trajectories with $n_p=5$ and $n_f=30$. To evaluate performance on unseen environments, we are using location ID $4$ only during testing time. The scene is labeled with the same $N_c = 5$ classes as in SDD.

\noindent \textbf{ETH \& UCY datasets}: The ETH/UCY benchmarks have been widely used for benchmarking trajectory forecasting models in the short horizon setting in the recent years \cite{PWCeth}. Forecasting performance has improved by over $\sim64\%$ on average, within the last two years itself \cite{gupta2018social}.

It comprises of five different scenes all of which report position in world coordinates (in meters). We follow the leave one out validation strategy as outlines in prior works \cite{gupta2018social, sadeghian2019sophie, deo2020trajectory, salzmann2020trajectron++}. We use the preprocessed data from \cite{gupta2018social} \footnote{https://github.com/agrimgupta92/sgan}, also used by state-of-the-art methods \cite{mangalam2020not, salzmann2020trajectron++}. Similar to SDD, the frames are sampled at $\texttt{FPS = 2.5}$ predicting $n_f = 12$ frames, $t_f = 4.8$ seconds into the future given last $n_p = 8$ frames comprising of $t_p = 3.2$ seconds of the past.

Our model represents trajectories as heatmaps and hence needs the coordinates in pixel space. The ETH/UCY data lie in world coordinates. To project the data from meter into pixel space we use the provided homography matrices from the original dataset for the ETH \cite{pellegrini2010improving} scenes ETH and HOTEL and create our own homography matrices for the UCY\cite{lerner2007crowds} scenes UNIV, ZARA1 and ZARA2. To enable fair comparisons, we convert our predictions back to world coordinates using the inverse homography matrices and calculate our errors with the untouched raw data in world coordinates, to avoid any errors from our projection. 

For all ETH and \& UCY datasets, since the classes of affordances furnished by the surfaces present is small, we use $N_c = 2$, identifying each pixel as either belonging to class `road' or `not road'. 

\begin{table*}[t!]
\begin{center}
\resizebox{\textwidth}{!}{
\begin{tabular}{c|c|c|c|c|c|c||c|c|c|c}
\toprule \midrule
     &   S-GAN  & CF-VAE & P2TIRL  & SimAug & PECNet* & $\textsf{Y}$-net (Ours)  & DESIRE  & TNT*  & PECNet  & $\textsf{Y}$-net (Ours)  \\  \hline\rule{0pt}{\normalbaselineskip}
    &\multicolumn{6}{c||}{$K = 20$} & \multicolumn{4}{c}{$K = 5$}  \\ \hline \rule{0pt}{\normalbaselineskip}ADE  &  27.23 & 12.60  & 12.58           & 10.27            & 9.96  & \textbf{7.85} & 19.25    & 12.23  &  12.79 & \textbf{11.49}  \\
FDE  &  41.44 & 22.30  & 22.07           & 19.71            & 15.88  &  \textbf{11.85}  & 34.05    & 21.16 & 29.58 & \textbf{20.23} \\ \bottomrule
\end{tabular}
}
\end{center}

\caption{\textbf{Short temporal horizon forecasting results on SDD}: Our method significantly outperforms previous state-of-the-art methods (indicated by *) on the Stanford Drone Dataset \cite{robicquet2016learning} on both the ADE \& FDE metrics for both settings of $K$, where $K$ represents the number of multimodal samples . Reported errors are in pixels with $t_p = 3.2 \text{ sec}, t_f = 4.8 \text{ sec}, n_p = 8, n_f = 12$. Lower is better.}

\label{tab:sdd_short}
\end{table*}

\begin{table*}[t!]
\begin{center}
\resizebox{\textwidth}{!}{
\begin{tabular}{c|c|c|c|c|c|c|c|c|c|c|c|c}
\toprule \midrule
& \multicolumn{2}{c|}{S-GAN} & \multicolumn{2}{c|}{Sophie}     & \multicolumn{2}{c|}{CGNS}       & \multicolumn{2}{c|}{PECNet} & \multicolumn{2}{c|}{Trajectron++*} & \multicolumn{2}{c}{$\textsf{Y}$-net (Ours)}      \\  \hline \rule{0pt}{\normalbaselineskip}
 & \multicolumn{1}{|l|}{ADE} & FDE  & \multicolumn{1}{l|}{\hspace{1mm} ADE \hspace{1mm}} & FDE  & \multicolumn{1}{l|}{ADE} & FDE  & \multicolumn{1}{l|}{ADE}  & FDE  & \multicolumn{1}{l|}{\ ADE\ \ } & FDE  & \multicolumn{1}{l|}{\ \ ADE\ \ \ } & FDE \\ \midrule
ETH                    & 0.81                     & 1.52 & 0.70                     & 1.43 & 0.62                     & 1.40 & 0.54                       & 0.87    & 0.39                     & 0.83  & \textbf{0.28}  &  \textbf{0.33}   \\ 
HOTEL                  & 0.72                     & 1.61 & 0.76                     & 1.67 & 0.70                     & 0.93 & 0.18 & 0.24   & 0.12                     & 0.21 & \textbf{0.10} & \textbf{0.14}      \\ 
UNIV                   & 0.60                     & 1.26 & 0.54                     & 1.24 & 0.48                     & 1.22 & 0.35                     & 0.60 &  \textbf{0.20}                     & 0.44 & 0.24 & \textbf{0.41}     \\ 
ZARA1 & 0.34 & 0.69 & 0.30 & 0.63 & 0.32   & 0.59 & 0.22   & 0.39  & \textbf{0.15} & 0.33 & 0.17 & \textbf{0.27} \\ 
ZARA2 & 0.42 & 0.84 & 0.38 & 0.78 & 0.35 & 0.71 & 0.17 & 0.30 & \textbf{0.11} & 0.25  & 0.13 & \textbf{0.22} \\ \hline \hline \rule{0pt}{\normalbaselineskip}AVG & 0.58 & 1.18 & 0.54 & 1.15 & 0.49 & 0.97 &  0.29  & 0.48  &  0.19 & 0.41 & \textbf{0.18} & \textbf{0.27}\\ \bottomrule
\end{tabular}
}
\end{center}

\caption{\textbf{Short term forecasting results on ETH/UCY benchmark}: Our proposed method establishes new state-of-the-art results (previous results denoted by *) on both the ADE \& the FDE metrics on the popular ETH-UCY benchmark using standard short-horizon settings (same as SDD) and $K=20$.  Reported errors are in meters. Lower is better. 
}

\label{tab:eth-ucy}
\end{table*}

\subsection{Implementation Details}
\subsubsection{Segmentation Model Implementation Details}
To incorporate constraints and interactions of the agents with the scene, we pretrain a semantic segmentation model to efficiently use the sparse scene image data. Stanford Drone Dataset contains 60 scene images in total, while inD only contains images from four recording locations. We use the U-net model \cite{unet} with ResNet101 \cite{he2016deep} backbone. The ResNet101 encoder's weights are pretrained on ImageNet, while the weights for the U-net decoder and segmentation head are randomly initialized. The images are downsampled by a factor of four (SDD) and three (inD), padded to be divisible by 32 as required for U-net and cropped to $256 \times 256$. The data is augmented spatially by rotation, flipping, scaling and perspective transformation and we introduce Gaussian noise, blurring as well as color, brightness and contrast shifts. The semantic maps are manually labeled into the $N_c=5$ classes mentioned above, as well as a dummy class for padding and black areas in the inD dataset. We only use the corresponding images from the trajectory train scenes for training to evaluate the performance of $\textsf{Y}$-net on unseen environments for both SDD and inD. 

The SDD segmentation model is trained using ADAM optimizer to reduce the Dice Loss \cite{sudre2017generalised} with an initial learning rate of $1 \times 10^{-4}$ and batch size of 4. The learning rate is decreased to $1 \times 10^{-5}$ after 1500 epochs. Further we freeze the ResNet101 backbone for the first 200 epochs. 

As inD only contains images from four locations, we use the pretrained SDD model and freeze the encoder for the first 1000 epochs to avoid catastrophic forgetting. All other hyper parameters are the same as for training SDD.

\begin{table*}[t!]
\begin{center}
\resizebox{\textwidth}{!}{
\begin{tabular}{@{}cccccccccc|ccc@{}}
\toprule \midrule
\multicolumn{1}{|l}{}     & \multicolumn{6}{c||}{Stanford Drone Dataset}                                                                                               & \multicolumn{6}{c|}{Intersection Drone Dataset}                                                     \\ \midrule
\multicolumn{1}{l|}{}    & \multicolumn{1}{c}{S-GAN} & \multicolumn{1}{c}{PECNet} & \multicolumn{1}{c|}{R-PECNet} & \multicolumn{3}{c||}{$\textsf{Y}$-net (Ours)}                                  & \multicolumn{1}{c}{S-GAN} & PECNet & \multicolumn{1}{c|}{R-PECNet} & \multicolumn{3}{c}{$\textsf{Y}$-net (Ours)}              \\ \hline 
\multicolumn{1}{c}{\rule{0pt}{\normalbaselineskip} $K_a$}   & \multicolumn{1}{|c}{1}                 & 1         & \multicolumn{1}{c|}{1}      & 1     & \multicolumn{1}{c}{2} & \multicolumn{1}{c||}{5}     & 1        & 1 & 1    & 1     & 2     & \multicolumn{1}{c}{5} \\ \hline 
\multicolumn{1}{c}{\rule{0pt}{\normalbaselineskip}ADE} &  \multicolumn{1}{|c}{155.32}             & 72.22       & \multicolumn{1}{c|}{261.27}  & 47.94 & 44.94                 & \multicolumn{1}{c||}{39.49} & \multicolumn{1}{c}{38.57}  & 20.25  & 341.80 & 14.99 & 14.02  & 12.67                  \\
\multicolumn{1}{c|}{FDE} & 307.88                   & 118.13  & \multicolumn{1}{c|}{750.42} & 66.71 & 66.71                 & \multicolumn{1}{l||}{66.71} & \multicolumn{1}{c}{84.61}  & 32.95  & 1702.64 & 21.13 & 21.13 & 21.13                     \\ \bottomrule
\end{tabular}
}
\end{center}

\caption{\textbf{Long term trajectory forecasting Results}: We benchmark performance on our proposed long horizon forecasting setting predicting $t_f = 30$ seconds into the future given $t_p = 5$ seconds past motion history. All reported error are in pixels (lower is better) for $K_e = 20$ with additional results for varying $K_a$ with a fixed $K_e$.}

\label{tab:sdd_long}
\end{table*}

\subsubsection{$\textsf{Y}$-net Implementation Details}
We train the entire network end to end with ADAM optimizer with a learning rate of $1 \times 10^{-4}$ and batch size of 8. We scale the overall loss by a factor of 1000.
Since the scene images $\mathcal{I}$ have different heights and widths in all datasets, we ensure that each batch only contains image and trajectories from the same scene. $\textsf{Y}$-net does not use fully-connected layers and therefore can handle images of different sizes, without cropping or padding to the same shape. The RGB scene image $\mathcal{I}$ and trajectory heatmaps $\mathbf{H}$ are downsampled by 4 for SDD, 3 for inD and 1.5 for ETH/UCY to save memory and padded to be divisible by 32. For fair comparisons with previous methods we upsample the predicted trajectories back to its original size and compare with the ground-truth data in original scale. All scene images and trajectories are augmented by spatial flipping and rotation in 90° steps, increasing the number of training data by a factor of eight. The encoder blocks in $\mathbf{U}_e$ have output channel dimensions $[32, 32, 64, 64, 64]$ and both $\mathbf{U}_g$ and $\mathbf{U}_t$ start with two convolutional layers of output channels $128$, followed by blocks of output channel dimensions $[64, 64, 64, 32, 32]$. We use $\lambda_1=\lambda_2=1$ to weight the binary cross entropy loss. The ground-truth trajectory heatmaps has a variance of $\sigma_H = 4$ pixels.

During training $\mathbf{U}_g$ predicts the goal and waypoint distribution for all $n_p$ time steps as an auxiliary task. This helps to let the sub-network learn the dynamics of pedestrian trajectories better. During inference, we only use the goal and $N^w$ waypoint distributions as needed.

The trajectory sub-network $\mathbf{U}_t$ is trained using the ground-truth goal and waypoints. Those are represented as trajectory heatmaps as described in Section \ref{sec:traj_on_scene_subsec} and downsampled spatially to fit the corresponding feature map shapes of $\mathbf{U}_t$ blocks. By using the ground-truth, $\mathbf{U}_t$ learns to predict trajectories leading towards the goal, while passing the waypoints. During inference, we use the ($\texttt{TTST}$) sampled goals and waypoints predicted by $\mathbf{U}_g$.

On ETH/UCY, we further experiment with deformable convolutional layers as proposed in \cite{dai2017deformable}.

We will release all our data, both raw \& processed, code for reproducing experiments across all the datasets \& labeled semantic maps for reproducibility and future work. 

\subsection{Metrics}
We use the established Average Displacement Error (ADE) and Final Displacement Error (FDE) metrics for measuring performance of future predictions. ADE is calculated as the $\ell_2$ error between the predicted future and the ground truth averaged over the entire trajectory while FDE is the $\ell_2$ error between the predicted future and ground truth for the final predicted point \cite{alahi2014socially}. Following prior works \cite{gupta2018social}, in the case of multiple future predictions, the final error is reported as the $\texttt{min}$ error over all predicted futures.

\subsection{Baseline models}
We benchmarks against several state-of-the-art methods across both short and long term trajectory forecasting settings which we describe briefly.  
\begin{itemize}
    \item Social GAN \cite{gupta2018social} proposes a Generative Adversarial Network (GAN) to predict multi-modal trajectory autoregressively.   
    \item SoPhie \cite{sadeghian2019sophie} also uses a GAN and extends it by attention modules to incorporate other agents and scene context.
    \item Conditional Flow VAE (CF-VAE) \cite{bhattacharyya2019conditional} proposes a conditional normalizing flow based Variational Auto-Encoder that models future uncertainty without disentangling underlying factors.
    \item Conditional Generative Neural System (CGNS) \cite{li2019conditional} proposes variational divergence minimization in latent space to learn feasible regions for future trajectories. 
    \item P2TIRL \cite{deo2020trajectory} proposes a grid based trajectory forecasting method learnt using maximum entropy inverse reinforcement learning. 
    \item DESIRE \cite{lee2017desire} also proposes an inverse reinforcement learning approach for prediction by planning.
    \item SimAug \cite{liang2020simaug} is a recently proposed method that uses additional adversarially generated 3D multi-view data for adapating to novel viewpoints in forecasting and improve the Multiverse model \cite{liang2020garden}.
    \item PECNet \cite{mangalam2020not} is the prior state-of-the art method on short-term trajectory prediction on the Stanford Drone Dataset. They propose to use goal-conditioning but does not account for multi-modality in the path to the goal.
    \item TNT \cite{zhao2020tnt} closely improves upon PECNet's performance for $K=5$ samples on SDD and is the prior state-of-the-art in that setting.
    \item Trajectron++ \cite{salzmann2020trajectron++} proposes a recurrent graph based forecasting model incorporating dynamic constrains such as other moving agents and scene information.  This work also hold the prior state-of-the art on ETH/UCY short-term trajectory prediction benchmark.
\end{itemize}

\subsection{Short Term Forecasting Results}

\subsubsection{Stanford Drone Results} Table \ref{tab:sdd_short} presents results on SDD in the short term setting \ie $t_p = 3.2$ seconds, $t_f = 4.8$ seconds. We follow the standard split from \cite{sadeghian2018trajnet} and report results with $K_e = 5 \ \& \ 20$. Since there's limited aleatoric multimodality in short term settings, we use $K_a = 1$ thus being comparable to prior works using $K=20$ trajectory samples for final evaluation. Table \ref{tab:sdd_short} shows our proposed model achieving an ADE of $7.85$ and FDE of $11.85$ at $K_e = 20$ which outperforms the previous state-of-the-art performance of PECNet \cite{mangalam2020not} by $26.8\%$ on ADE and $34.0 \%$ on FDE. Further, at $K =5$ it achieves an ADE of $11.49$ \& FDE of $20.23$ outperforming previous state-of-the-art performance of TNT. 

\subsubsection{ETH \& UCY Results} We also report results on the ETH/UCY benchmark in Table \ref{tab:eth-ucy}. Similar to SDD, we set $K_e = 20, K_a = 1$. Here as well, we observe that our proposed model achieves an ADE of $0.18$ \& FDE of $0.27$ improving the previously state-of-the-art performance from Trajectron++ \cite{salzmann2020trajectron++} of $0.19$ ADE \& $0.41$ FDE by about $5.6\%$ ADE and $51.9\%$ FDE. 

\subsection{Long Term Forecasting Results}
To study the effect of \textit{epistemic} \& \textit{aleatoric} uncertainty, we propose a long term trajectory forecasting setting with a prediction horizon upto $10$ times longer than prior works, extending to a minute. To benchmark, we retrain PECNet \cite{mangalam2020not}, the previous state-of-the-art method from short term forecasting \& Social GAN \cite{gupta2018social} for each $t_f$ in the long horizon setting. We also train a recurrent short term baseline using the PECNet model (R-PECNet) where the model is trained only for $t_f = 5$ seconds and is fed its own predictions recurrently for predicting longer temporal horizons. 

\begin{table}[]
\begin{center}
\resizebox{0.48\textwidth}{!}{
\begin{tabular}{@{}ccccc|cc@{}}
\toprule \midrule
\multicolumn{1}{c|}{}     & \multicolumn{3}{c||}{SDD}                & \multicolumn{2}{c}{inD}                                                     \\ \midrule
\multicolumn{1}{c|}{$\texttt{TTST}$}    & \multicolumn{1}{c}{\ding{55}} & \ding{55} &  \multicolumn{1}{c||}{\ding{51}}                                  & \multicolumn{1}{c}{\ding{55}} & \ding{51}       \\
\multicolumn{1}{c|}{$\texttt{CWS}$}    & \multicolumn{1}{c}{\ding{55}} & \ding{51} &  \multicolumn{1}{c||}{\ding{51}}                                  & \multicolumn{1}{c}{\ding{51}} & \ding{51}        \\ \hline
\multicolumn{1}{c|}{\rule{0pt}{\normalbaselineskip}ADE} & \multicolumn{1}{c}{65.00}  &      52.31   & \multicolumn{1}{c||}{47.94} & \multicolumn{1}{c}{17.77}  & 14.99                  \\
\multicolumn{1}{c|}{FDE} & \multicolumn{1}{|c}{86.98} &\multicolumn{1}{c}{86.98}         & \multicolumn{1}{c||}{66.71} & \multicolumn{1}{c}{28.52}  & 21.13                     \\ \bottomrule
\end{tabular}
}
\end{center}

\caption{\textbf{Ablation results for Conditioned Waypoint Sampling ($\texttt{CWS}$) and $\texttt{TTST}$}: We benchmark the performance of $\textsf{Y}$-net with and without our proposed $\texttt{CWS}$ and $\texttt{TTST}$ on our long horizon forecasting setting, predicting $t_f = 30$ seconds into the future given $t_p = 5$ seconds of past motion history. All reported errors are in pixels (lower is better) for $N^w=1$, $K_e = 20$ and $K_a=1$.}

\label{tab:ttst}
\end{table}

\subsubsection{Forecasting Results}Table \ref{tab:sdd_long} reports the baseline and our results on the Stanford Drone (SDD) and Intersection Drone Datasets (InD) for a time horizon of $t_f = 30$ seconds in the future given the past $t_p = 5$ seconds input. All reported results are with $K_e = 20$ for $\textsf{Y}$-net conditioned on $N_w = 1$ intermediate waypoint at $w_i = 20$, \ie midway temporally between the observed inputs and the estimated goal. All reported baseline results are at $K=20$ for fair comparisons with our $K_e = 20, K_a = 1$ setting. On SDD, we observe that our proposed model outperforms the state-of-the-art short term baseline on the long horizon setting as well, achieving an ADE of $47.94$ and FDE $66.72$ improving upon PECNet's performance by over $50\%$. Similarly, $\textsf{Y}$-net outperforms PECNet on InD improving ADE performance from $20.25$ to $14.99$ and FDE from $32.95$ to $21.13$. 

\begin{figure}

\begin{center}
\includegraphics[width = \linewidth]{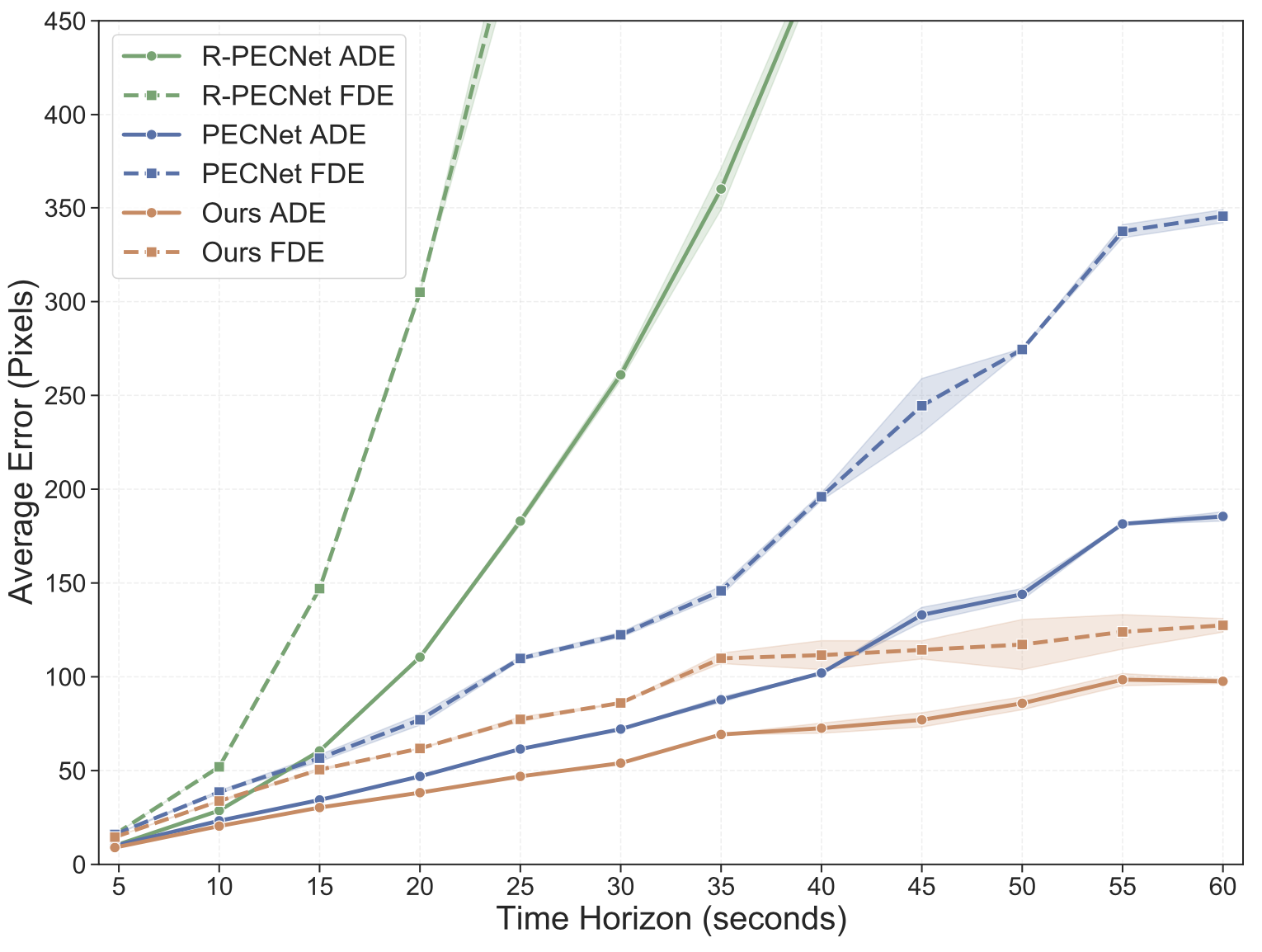}
\end{center}
  \caption{\textbf{Benchmarking Performance against Time Horizons}: On prediction horizons upto a minute, we observe a consistently growing difference in ADE between $\textsf{Y}$-net and PECNet, highlighting the important of factorized goal \& path modeling in long term forecasting.}
\label{fig:time}

\end{figure}

\begin{figure}
    \centering
    \includegraphics[width=\linewidth]{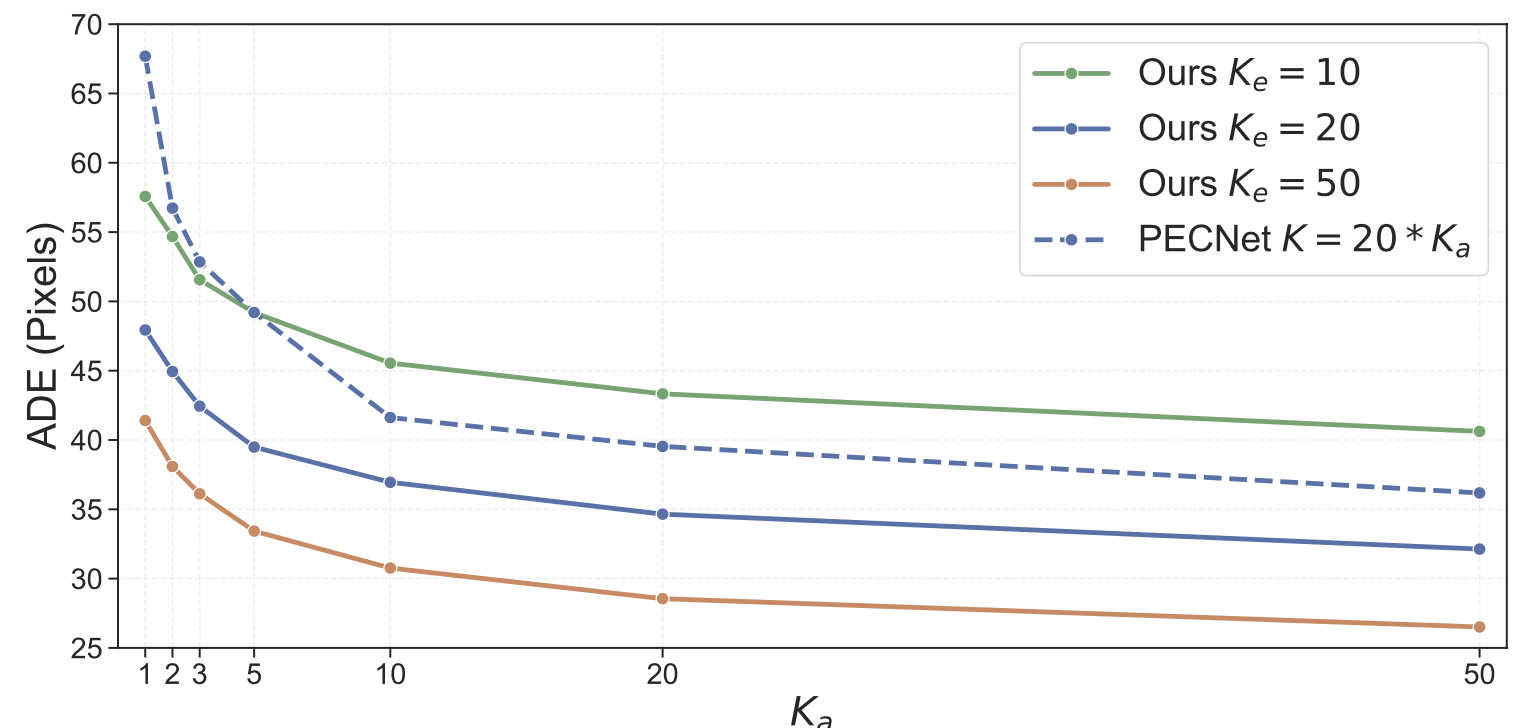}

    \caption{\small{\textbf{Benchmarking performance against aleatoric uncertainty} ($K_a$): Fixing the goal multimodality ($K_e$) we vary $K_a$ to observe the effect of path multimodality. Also, we benchmark against PECNet by allowing it $20$ times more samples for each $K_a$ for a fair compare against the $K_e = 20$ $\textsf{Y}$-net curve.}}
    \label{fig:vary_k_e}

\end{figure}

\begin{figure*}[t!]
\centering
\begin{subfigure}{\textwidth}
    \centering
    \includegraphics[trim={1cm 0 0 0}, clip, angle=180, width=0.24\textwidth]{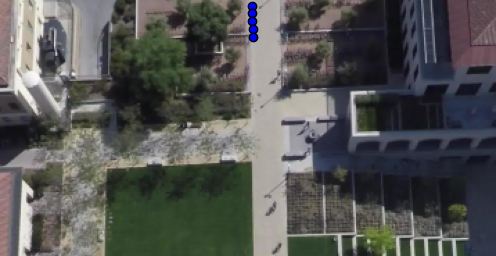}
    \includegraphics[trim={1cm 0 0 0}, clip, angle=180, width=0.24\textwidth]{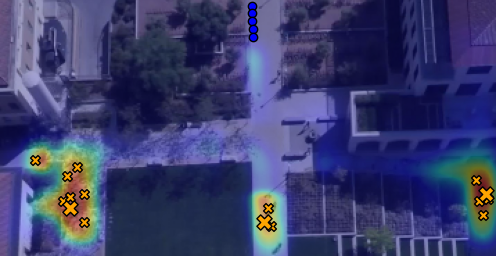}
    \includegraphics[trim={1cm 0 0 0}, clip, angle=180, width=0.24\textwidth]{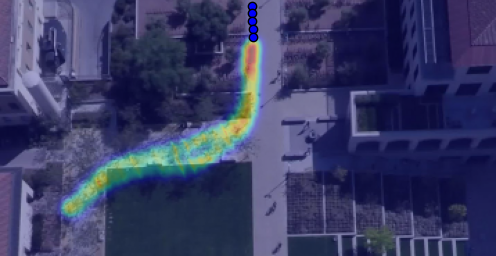}
    \includegraphics[trim={1cm 0 0 0}, clip, angle=180, width=0.24\textwidth]{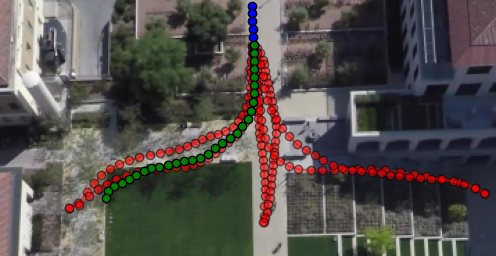}
\end{subfigure}
\begin{subfigure}{\textwidth}
    \centering
    \includegraphics[height=0.24\textwidth, angle=90]{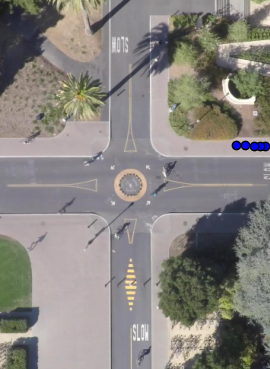}
    \includegraphics[height=0.24\textwidth, angle=90]{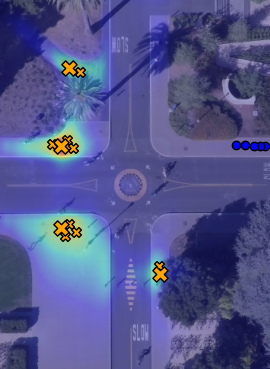}
    \includegraphics[height=0.24\textwidth, angle=90]{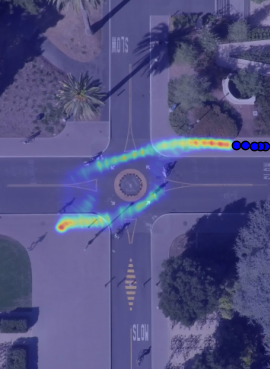}
    \includegraphics[height=0.24\textwidth, angle=90]{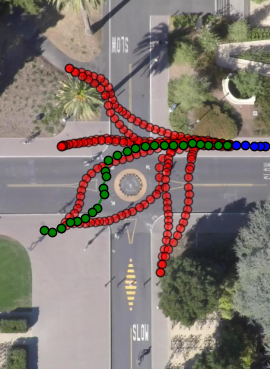}
\end{subfigure}
\begin{subfigure}{\textwidth}
    \centering
    \includegraphics[height=0.24\textwidth, angle=90]{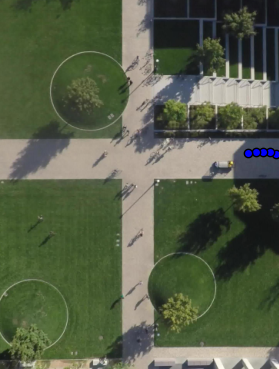}
    \includegraphics[height=0.24\textwidth, angle=90]{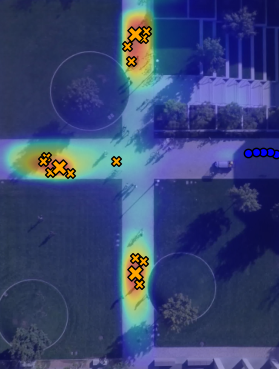}
    \includegraphics[height=0.24\textwidth, angle=90]{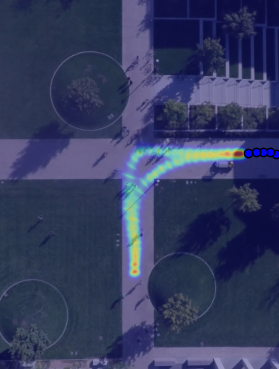}
    \includegraphics[height=0.24\textwidth, angle=90]{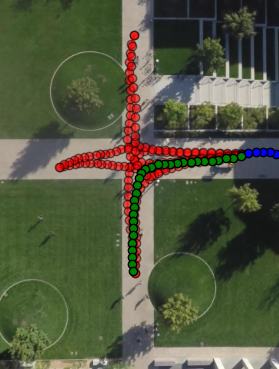}
\end{subfigure}

\caption[short]{\textbf{Qualitative Long Term Trajectory Forecasting Results}: We show various heatmaps and visualizations for three different scenes (rows) in SDD testset. The first column shows the past observed trajectory for last $t_p =5$ seconds in blue. The second column shows the heatmap from $\mathbf{U}_g$ for $t_f = 30$ seconds in the future (goal multimodality) and some sampled goals from the estimated distribution. The third column shows trajectory heatmaps from $\mathbf{U}_t$ conditioned on a sampled goal from column three (path multimodality). The last column shows the predicted trajectories, green indicating the ground-truth trajectories \& red our multimodal predictions.}
\label{fig:qual}
\end{figure*}

\subsubsection{Conditoned Waypoint Sampling}
Table \ref{tab:ttst} shows an ablation of the Conditioned Waypoint Sampling $\texttt{CWS}$. While it doesn't affect the goal sampling, \ie the FDE, the ADE decreases by $24.3\%$.

\subsubsection{Test-Time Sampling Trick}
Table \ref{tab:ttst} shows the effectiveness of our proposed $\texttt{TTST}$. $\texttt{TTST}$ reduces the error on SDD and inD by $9.1\%$ and $18.5\%$ in ADE, respectively, and $30.4\%$ and $35.0\%$ in FDE. 

\subsubsection{Varying Prediction Horizon} We also compare $\textsf{Y}$-net with two flavours of PECNet, one retrained separately for each prediction horizon $t_f$, and another trained only for $t_f = 5$ seconds but evaluated recurrently for long horizons (R-PECNet). We observe that the difference in ADE between $\textsf{Y}$-net \& PECNet grows as prediction horizon increases from $5$ to $60$ seconds. This shows $\textsf{Y}$-net's adaptability for long prediction horizons owing to factorized mulitmodality modeling. We also observe that for PECNet, training a separate model for different time horizons is significantly better than using a short temporal horizon model recurrently. This motivates our proposal for studying long term forecasting since short term models behave very poorly when applied out of box recurrently to longer term settings.

\subsubsection{Varying $K_a$} We also report results with $K_a = 2 \ \& \ 5$ for studying the improvement in performance from aleatoric multimodality in Table \ref{tab:sdd_long}. We observe a consistent improvement in ADE on both datasets, thus indicating the diversity in predicted paths given the same estimated final goal $\mathbf{u}_{n_p + n_f}$. We also report extensive results for varying the path multimodality $K_a$ with a fixed $K_e$ for various choice of $K_e$ \& $K_a$ in Figure \ref{fig:vary_k_e}. Additionally for baselining, we benchmark against PECNet \cite{mangalam2020disentangling} evaluated with $K_e$ times more samples than the corresponding $\textsf{Y}$-net model while varying $K_a$. We show consistent ADE improvements for various $K_e$ while increasing $K_a$, indicating effective use of multimodality. Further, even with $K = K_e*20$ samples, \ie 20 times more samples for each $K_e$, PECNet's performance is significantly worse than $\textsf{Y}$-net at $K_e = 20$ for all $K_a$ highlighting the importance of factorizing goal and path multimodality for diverse \& accurate future trajectory modeling. 

\begin{figure}
    \animategraphics[loop, autoplay, width=0.5\textwidth]{2}{figures/gif/frame_}{0}{44}
    \animategraphics[loop, autoplay, width=0.5\textwidth]{2}{figures/gif2/frame_}{0}{44}
    \caption{\small{\textbf{GIF Visualization}: Demonstrating the goal, waypoint and path multimodality for long term human trajectory prediction ($30$ seconds horizon). Given the past $5$ seconds input history (green), we predict diverse future trajectories (current location in orange, past in red). \textit{Best viewed in Adobe Acrobat Reader}.}}
    \label{fig:gif}
    \vspace{5mm}
\end{figure}

\subsubsection{Qualitative Results}
We show qualitative results for long term trajectory prediction ($t_f = 30$) on SDD in Figure \ref{fig:qual} and through a GIF temporally in Figure \ref{fig:gif}. We observe that $\textsf{Y}$-net predicts diverse scene-complaint trajectories, with both future goal and path multimodalities.

\section{Conclusion}
In summary, we present $\textsf{Y}$-net, a scene-compliant trajectory forecasting network with factorized goal and path multimodalities. $\textsf{Y}$-net uses the U-net structure \cite{unet} for explicitly modeling probability heatmaps for epistemic and aleatoric uncertainties. Overall, $\textsf{Y}$-net improves previous state-of-the-art performance by $34.0\%$ on the SDD and by $51.9\%$ on ETH/UCY benchmarks in the short term setting. We also propose a new long term trajectory forecasting setting (prediction horizon upto a minute) for exemplifying the epistemic and aleatoric uncertainty dichotomy. In this setting, we benchmark on the Stanford Drone \& Intersection Drone dataset where $\textsf{Y}$-net exceeds previous state-of-the-art by over $77.1\%$ and $55.9\%$ respectively thereby highlighting the importance of modeling factorized stochasticity.

{\small
\bibliographystyle{ieee_fullname}
\bibliography{egbib}
}
\end{document}